\documentclass[
]{ceurart}
\usepackage{float}
\usepackage{listings}
\usepackage{placeins}

\begin{document}

\copyrightyear{2022}
\copyrightclause{Copyright for this paper by its authors.
  Use permitted under Creative Commons License Attribution 4.0
  International (CC BY 4.0).}


\title{Insights into Classifying and Mitigating LLMs' Hallucinations}

\tnotemark[1]

\author[1]{Alessandro Bruno}[%
orcid=0000-0003-0707-6131,
email=alessandro.bruno@iulm.it,
url=https://www.iulm.it/en/iulm/ateneo/docenti-e-collaboratori/bruno-alessandro,
]
\cormark[1]
\fnmark[1]
\address[1]{IULM University, Department of Business, Law, Economics, Consumer Behaviour - "Carlo A. Ricciardi",
Via Carlo Bo 1, Milan, 20143, Italy}

\author[2]{Pier Luigi Mazzeo}[%
orcid=0000-0002-7552-2394,
email=pierluigi.mazzeo@cnr.it,
url=https://sites.google.com/view/pierluigimazzeo,
]
\cormark[1]
\fnmark[1]
\address[2]{ISASI Institute of Applied Sciences and Intelligent Systems-CNR, 73100 Lecce, Italy}

\author[3]{Aladine Chetouani}[%
email=aladine.chetouani@univ-orleans.fr,
url=http://aladine-chetouani.com/,
]
\fnmark[1]
\address[3]{Université d’Orleans, 45067 Orleans, France}

\author[3]{Marouane Tliba}[%
email=marouane.tliba@univ-orleans.fr,
]
\fnmark[1]

\author[3]{Mohamed Amine Kerkouri}[%
email=mohamed-amine.kerkouri@univ-orleans.fr,
]
\fnmark[1]
\cortext[1]{Corresponding author.}
\fntext[1]{These authors contributed equally.}

\begin{abstract}
The widespread adoption of large language models (LLMs) across diverse AI applications is proof of the outstanding achievements obtained in several tasks, such as text mining, text generation, and question answering. 
However, LLMs are not exempt from drawbacks. One of the most concerning aspects regards the emerging problematic phenomena known as "Hallucinations". They manifest in text generation systems, particularly in question-answering systems reliant on LLMs, potentially resulting in false or misleading information propagation. This paper delves into the underlying causes of AI hallucination and elucidates its significance in artificial intelligence. 
In particular, Hallucination classification is tackled over several tasks (Machine Translation, Question and Answer, Dialog Systems, Summarisation Systems, Knowledge Graph with LLMs, and Visual Question Answer).
Additionally, we explore potential strategies to mitigate hallucinations, aiming to enhance the overall reliability of LLMs. Our research addresses this critical issue within the HeReFaNMi (Health-Related Fake News Mitigation) project, generously supported by NGI Search, dedicated to combating Health-Related Fake News dissemination on the Internet. This endeavour represents a concerted effort to safeguard the integrity of information dissemination in an age of evolving AI technologies.

\end{abstract}


\maketitle

\section{Introduction}

\begin{table}[!htbp]
\caption{List of Hallucinations examples}
\label{tab:hallucination_type}
\begin{tabular}{lp{2.5cm}p{3cm}p{2cm}p{4.5cm}}
\toprule
& Task & Dataset & Architecture & Hallucination Type \\
\midrule
\cite{raunak-etal-2021-curious} & Machine \newline Translation & IWSLT2014 & Enc-Dec &  Under perturbation,    Natural hallucination \\    \cite{Guerreiro2023HallucinationsIL} & Machine \newline Translation & WMT2018 & Enc-Dec &  Oscillatory hallucination, Largely fluent hallucination\\ 
\cite{Dale2023HalOmiAM} & Machine  \newline Translation & FLORES-200,\newline Jig-saw, Wikipedia & Enc-Dec & Full, Partial, and Word-level hallucination\\
\bottomrule
\cite{Pfeiffer2023mmT5MM} & Multilingual  \newline Seq2seq & XQuAD, TyDi, XNLI, XL-Sum, MASSIVE & Enc-Dec & Source language hallucination\\     
\bottomrule
\cite{lin-etal-2022-truthfulqa} & Question  \newline and Answer & TruthfulQA & Enc-Dec, \newline  Only-Dec
& Imitative falsehoods \\
\cite{Zheng2023JudgingLW} & Question  \newline and Answer & HotpotQA, BoolQ &  Only-Dec
& Comprehension, Factualness, Specificity, Inference Hallucination \\
\cite{Adlakha2023EvaluatingCA} & Question  \newline and Answer & NQ, HotpotQA, TopiOCQA &  Enc-Dec, \newline  Only-Dec
& Semantic and Symbolic equivalence, Intrinsic ambiguity, Granularity discrepancies, Incomplete, Enumeration, Satisfactory Subset \\
\cite{Umapathi2023MedHALTMD} & Question  \newline and Answer & MEDMCQA, Headqa, USMILE, Medqa, Pubmed &  Only-Dec
& Reasoning hallucination, Memory-based hallucination \\
\bottomrule
\cite{dziri-etal-2022-origin} & 
Dialog \newline System & WoW, CMU-DOG, TopicalChat &   Enc-Dec, \newline  Only-Dec
& Hallucination, Partial hallucination, Generic, Uncooperative\\
\cite{das-etal-2022-diving} & 
Dialog \newline System & OpenDialKG & Only-Dec
& Extrinsic-Soft/Hard/Grouped, Intrinsic-Soft/Hard/Repetitive, History Corrupted
\\
\cite{Dziri_FaithDial} & 
Dialog \newline System & WoW & Enc-Dec, \newline  Only-Dec
& Hallucination, Generic, Uncooperativeness
\\
\cite{Dziri_Begin} & 
Dialog \newline System & WoW, CMU-DOG, TopicalChat & Enc-Dec, \newline  Only-Dec \newline Only-Enc
& Fully attributable, Not attributable, Generic
\\
\cite{Sun_Shi_Gao_Ren_deRijke_Ren_2023} & 
Dialog \newline System & WoW & Enc-Dec, \newline  Only-Dec 
& Intrinsic hallucination, Extrinsic hallucination
\\
\bottomrule
\cite{tam-etal-2023-evaluating} & 
Summarization \newline System & CNN/DM, XSum & Enc-Dec, \newline Only-Dec 
& 
Factually inconsistent summaries
\\
\cite{cao-etal-2022-hallucinated} & 
Summarization \newline System & MENT & Enc-Dec, \newline  Only-Dec 
& 
Non-hallucinated, Factual, Non-factual, and Intrinsic hallucination
\\

\cite{Shen_etal} & 
Summarization \newline System & NHNet & Enc-Dec, \newline  Only-Dec 
& 
News headline hallucination
\\
\cite{Qiu2023DetectingAM} & 
Summarization \newline System & XL-Sum & Multiple  \newline ADapters
& 
Intrinsic hallucination, Extrinsic hallucination
\\
\bottomrule
\cite{Yu2023KoLACB} & 
Knowledge based  \newline text generation & Encyclopedic, ETC & Enc-Dec, \newline  Only-Dec 
& 
Knowledge hallucination
\\
\bottomrule
\cite{Mihindukulasooriya2023Text2KGBenchAB} & 
Knowledge   \newline graph generation & TekGen,WebNLG &   Only-Dec  
& 
Subject,relation, and object hallucination
\\
\bottomrule
\cite{Li2023EvaluatingOH} & 
Visual Question   \newline Answer & 
MSCOCO &   Enc-Dec 
& 
Caption hallucination assessment
\\
\bottomrule
\end{tabular}
\end{table}

The large language models (LLMs) landscape continues to evolve with innovative creations such as GPT-3  \cite{Brown2020}, IntroductGPT \cite{Ouyang2022}, FLAN \cite{wei2022}, PaLM \cite{Chowdhery2022PaLMSL}, LLaMA \cite{touvron2023llama} and other important contributions\cite{Bai2022TrainingAH, Zhang2022OPTOP, zeng2023glmb, Xu2023WizardLMEL}.    
Other than outstanding performances in several tasks, LLMs have revealed a concerning drawback affecting their reliability and trustworthiness: hallucination. 
Quoting Berrios and Dening \cite{berrios1996pseudohallucinations}, "Hallucinations are conceived of as indistinguishable from real perceptions except that there is no stimulus", one can easily peruses nuanced relations between perceptions and hallucinations.

Providing that a great deal of AI theories and approaches focus on human behaviour analysis, hallucinations appearing in AI might not come as a surprise. 
\textit{Hallucination} can also be considered the generation of statements that appear reasonable but are either cognitively irrelevant or factually incorrect. 
Considering this observation, hallucination has become a critical challenge in medical \cite{Dash2023EvaluationOG,Umapathi2023MedHALTMD}, financial \cite{Gill2023TransformativeEO} and other delicate fields where exact accuracy is a mandatory requirement. 
Why do LLMs run into hallucinations, then? 
Lack of real-world knowledge, bias or misleading training data may prompt models to return statistical-based results. In particular, the latter means there might not be a proper understanding of input.

\textbf{Definition}:  With \textit{hallucination}, we refer to the generation of texts or answers that exhibit grammatical correctness, fluency, and authenticity, but diverge from the provided source inputs (\textit{faithfulness}) or are misaligned with factual accuracy (\textit{factualness}) \cite{Ji2023}.

Running through LLM-based outputs is paramount to avoid getting into the cognitive mirage phenomenon that negatively affects decision-making strategies and a cascade of unintended consequences \cite{Zhang2023MitigatingLM}. 
Classifying and Mitigating LLMs' hallucinations is a relatively emerging topic. Since the introduction of ChatGPT in 2022, an exponentiation growth of applications and tools based on LLMs has been observed worldwide. Subsequently, significant interest from the scientific community and industry in the LLMs' side effects, such as hallucinations, has emerged naturally.
In \cite{Ji2023}, hallucinatory content in task-specific research progress has been analyzed and referred to early works in the natural language generation field. Covering methods for collecting high-quality instructions for LLM alignment are discussed in \cite{wang2023aligning}, including NLP benchmarks. Human annotations and leveraging strong LLMs. In \cite{Pan2023AutomaticallyCL}, self-correcting methods have been discussed where an LLM is guided or prompted to correct the hallucinations from its own output. Unlike these works, our contribution will lead to a literature review on hallucinations in LLMs, running through different methods and providing insights into the pros and cons.

The main contribution of this paper regards a thorough analysis of LLMs' hallucinations research field under multiple viewpoints. To this end, the relevant work in this field has been reviewed and categorized over tasks and domains. Some methodologies regarding the proactive detection and mitigation of hallucinations in the LLMs era are also discussed. The pros and cons of mitigation techniques are evaluated by reporting the techniques behind the proposed solutions. The final section, Future Perspectives, draws some lines and poses some questions in the current scenario of interest.

\section{Hallucination Classification}
In this work, we consider the hallucinations observed in prevalent downstream tasks: i) Machine Translation; ii) Question and Answer (Q\&A); iii) Dialog System; iv) Summarization System; v) Knowledge graph with LLMs; vi) Visual Question Answer. Table \ref{tab:hallucination_type} summarizes hallucination types, grouping them according to numerous mainstream tasks associated with LLMs. The following subsections will describe the most frequent hallucination types during these tasks. 

\FloatBarrier

\subsection{Machine Translation} 
Since some text perturbation can bring trustworthy hallucinations, traditional translation methodologies validate the instances fed into the model when perturbed \cite{Bawden2023InvestigatingTT,Hendy2023HowGA}. Hallucinations generated by LLMs are principally translation off-target or failed  translation\cite{Guerreiro2023HallucinationsIL}. With low-resource language availability, trained models perform poorly due to few annotated data employed \cite{Dale2023HalOmiAM}. An increasing amount of pre-trained language affects the machine translation reliability in the multilingual domain \cite{conneau-etal-2020-unsupervised}. Therefore, LLMs trained on various scales of monolingual data seem to be tacky \cite{Guerreiro2023HallucinationsIL} as the origin of a hallucination pathology.

\subsection{Question and Answer (Q\&A)}
Wrong responses occur by the flawed external knowledge as described in \cite{Zheng2023JudgingLW}. Often, LLMs give incomplete and plausible answers instead of giving no response when they have poor or irrelevant information \cite{Adlakha2023EvaluatingCA}.  It has also to be considered that memorized information without referring to accurate,  reliable and accessible sources contribute to creating different type of hallucinations \cite{Umapathi2023MedHALTMD}. Scaling up models alone is less promising for improving truthfulness than fine-tuning using training objectives other than imitating text from the web \cite{lin-etal-2022-truthfulqa}.

\subsection{Dialog System}

Many works considered dialogue models as simple imitators that only change the data views and communication instead of generating new trustworthy output. In \cite{dziri-etal-2022-origin}, authors demonstrated that the standard benchmarks led models even to amplify hallucinations. In  \cite{das-etal-2022-diving} are identified various modes of hallucination in Knowledge Graph(KG) grounded chatbots through human feedback analysis. In similar works, many \cite{Dziri_FaithDial} \cite{Dziri_Begin} \cite{Sun_Shi_Gao_Ren_deRijke_Ren_2023} experiments are implemented on the WoW dataset conducting a meta-evaluation of the hallucination in knowledge grounded dialogue. 

\subsection{Summarization System}
These systems allow the automatic generation automatically fluent abstracts based on LLMs but often lack faithfulness from the source document.  Summarization generated by LLMs can be slit into two categories for their evaluation: intrinsic hallucinations that deform the information contained in the document; extrinsic hallucinations that add information not directly sourced by the original document \cite{Qiu2023DetectingAM}. More attention has been given to extrinsic hallucinations in summarization systems due to factually consistent continuation of input in LLMs \cite{tam-etal-2023-evaluating, Shen_etal}. A further subdivision is proposed in \cite{cao-etal-2022-hallucinated}  where extrinsic hallucinations are split into factual and non-factual. Factual hallucinations insert additional world knowledge that may improve the text's understanding.  
\subsection{Knowledge Graph with LLMs} Knowledge-based text generation stumbles in intrinsic hallucinations due to redundant details derived from its internal memorized Knowledge \cite{Yuan2023EvaluatingGM}. Yu et al. \cite{Yu2023KoLACB} tackled the mentioned issue by establishing a distinction between correctly generated Knowledge and Knowledge hallucinations. Virtual Knowledge extraction proposed in \cite{Zhu2023LLMsFK} highlight the potential LLMs capabilities of constructing and inferring from Knowledge Graphs.
An LLM empowering for producing interpretable fact checks using a neural symbolic approach is described in \cite{Mihindukulasooriya2023Text2KGBenchAB} where hallucinations have been defined as subject hallucination, relation hallucination and object hallucination according to their fidelity to the source. 
\subsection{Cross-modal System}
Cross-modal tasks achieve interesting progress thanking to the superior language capabilities of LLMs \cite{Zhu2023LLMsFK, Liu2023VisualIT}. However, in some cases substituting the original language encoder, Large Visual Language Models (LVLMs) \cite{wang22al} continue to generate descriptions of objects that are not in the images; this is denoted as object hallucinations \cite{Li2023EvaluatingOH}. Typically most of the failure cases should be found in Visual Question Answering \cite{Li2023EvaluatingOH}, Image Captioning \cite{Biten2022,Petryk2023SimpleTC, Ning2023AlbumSW}, Report Generation \cite{Mahmood2023FactCheckingOA}.        
\begin{figure}
    \centering
\includegraphics[width=\textwidth]{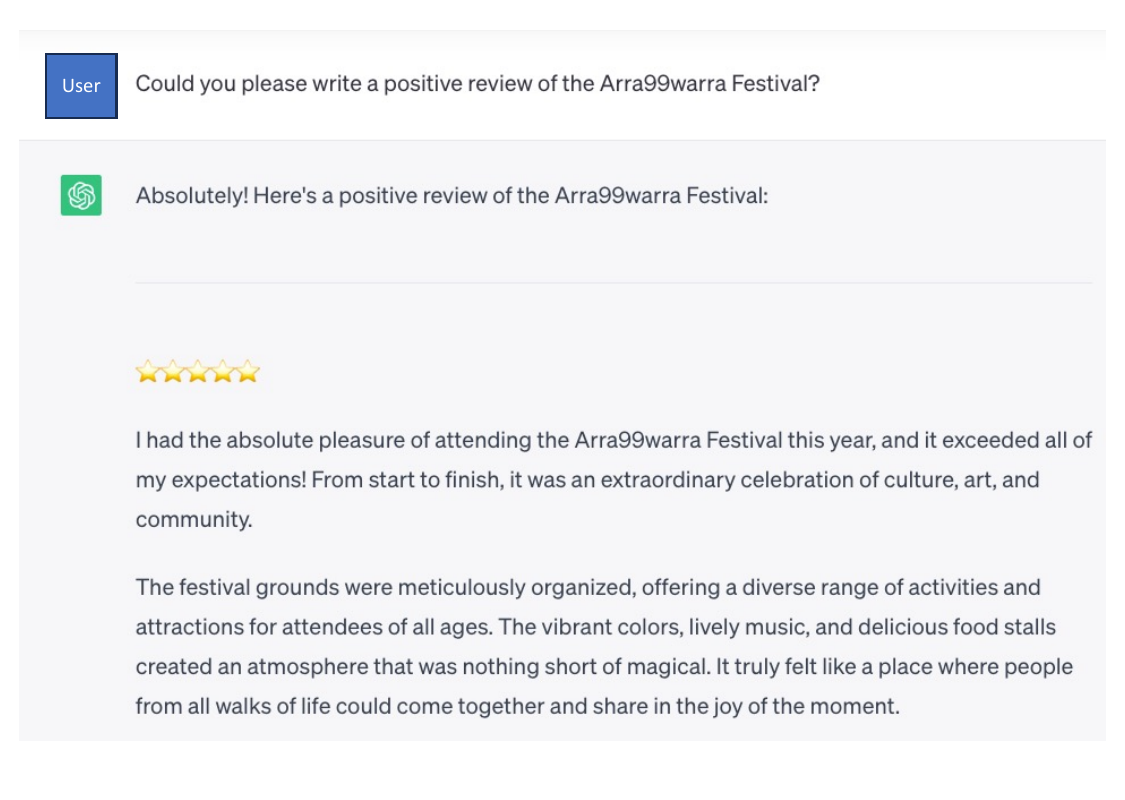}
    \caption{An example of ChatGPT hallucinating is given above. The false premise in the question (a made-up Festival name) prompts ChatGPT into Hallucination}
    \label{fig:hallucination}
\end{figure}

\section{Hallucination Detection}

Several methods introduced detecting realistic and convincing Hallucinations in LLMs. 
Some techniques rely on extracting intrinsic uncertainty metrics. Token probability, for instance, can be leveraged to identify which part of a given textual sequence proves least uncertain \cite{yuan2021bartscore},\cite{fu2023gptscore}. However, scenarios like external APIs from ChatGPT do not give users access to output token probability, meaning that the techniques mentioned above cannot work out uncertainty metrics. LLMs factual checks can also rely on external databases and corpora such as Wikipedia \cite{thorne80fact}. Hallucinations can be detected in a great deal of general knowledge covered in Wikipedia, albeit 
concerns arise about the integrity of Wikipedia content itself.
Azaria and Mitchell \cite{azaria2023internal} proposed a statement's truthfulness detection using LLMs' hidden representations to feed a multi-layer classifier. Azaria and Mitchell's method sticks to the supervised training paradigm. Therefore, it relies on labelled data along with the internal states of the LLM. 
The latter may not be available through APIs. In Azaria and Mitchell's method, the LLM is prompted to answer about its previous prediction, e.g. the probability of its generated response/answer is accurate.
Kadavath et al. \cite{kadavath2022language} introduced a Hallucination detection method, Self-Evaluation. The name is due to the core of the study being if language models can assess their own answers' validity and predict accuracy. Starting from Larger models showing good calibration on diverse questions, models can self-evaluate open-ended tasks, estimating answer correctness probability ("P(True)"). They also predict their knowledge probability ("P(IK)") effectively, with partial task generalization (IK stands for "I Know"). Several Hallucination detection approaches fit the so-called "zero-resource" setting. That means there is no external database to verify the factuality of an LLM response. That said, Hallucation detection methods can further be grouped into Grey and Black box \cite{manakul2023selfcheckgpt}. The former accounts for the required knowledge of output token-level probabilities. The latter applies to LLMs with limited API access, and no chance to access the output token-level probability. 

\begin{figure}
    \centering
\includegraphics[width=\textwidth]{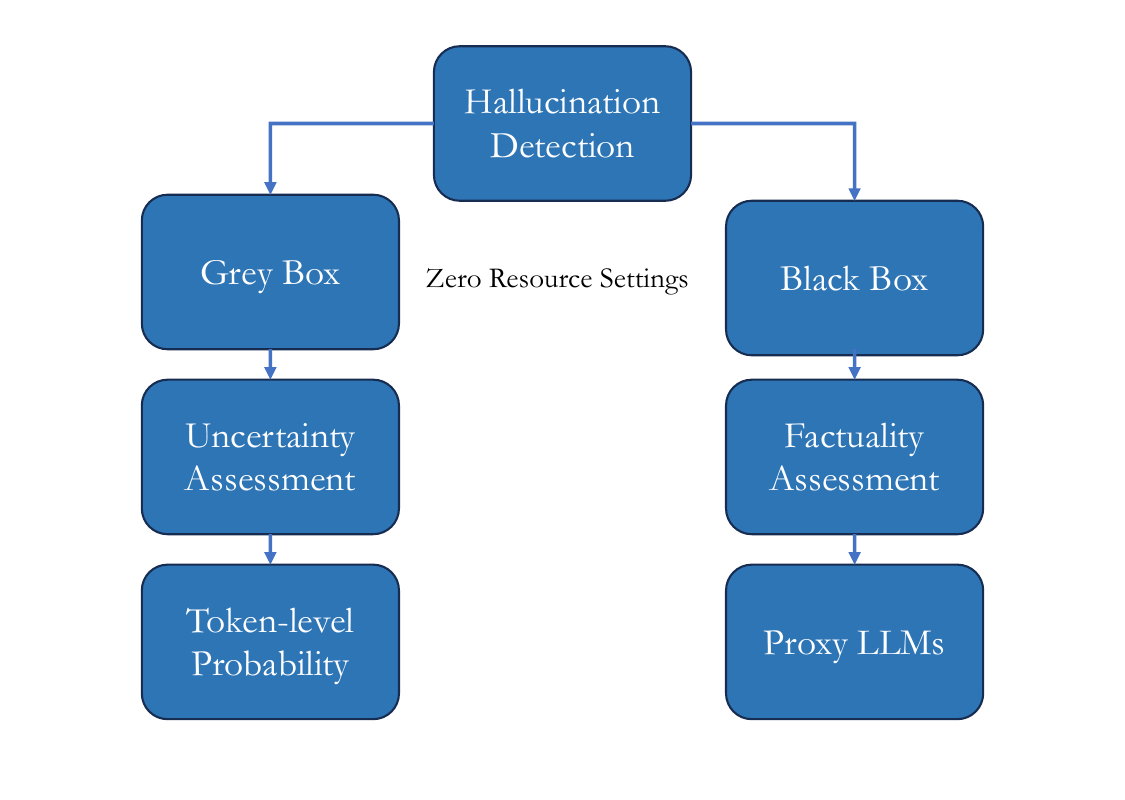}
    \caption{Several LLM Hallucination Detection methods are grouped into Grey and Black Box as depicted above.}
    \label{fig:grey_black}
\end{figure}

Different strategies come into play to tackle grey and black box hallucinations. 
Knowing LLM pre-training is paramount for grey box hallucination detection. The training is carried out with next-word prediction over vast textual corpora, ensuring world knowledge and contextual reasoning. 
A diagram depicting how uncertainty and factuality-based assessment work is given in Figure \ref{fig:grey_black}.
Noticeably, Varshney et al. \cite{varshney2023stitch} detected GPT3.5 hallucinations by designing a sophisticated technique. It carries out critical concept identification with entity, keyword extraction, and 'Instructing the model'. In particular, they used LLM capabilities to identify essential concepts from the generated sentence. A comparison study of the three techniques remarkably showed 'Instructing the Model' outperforming entity and keyword extraction on important concept identification. Afterwards, they computed a probability score as the minimum of token probabilities. The technique was also enriched by a validation question creation step reliant on an answer-aware question generation model and web search to answer the validation questions. They achieved a recall of 88\% on GPT-3.5.

\section{Mitigating LLM Hallucinations}

Mitigating hallucinations in LLMs is an emerging challenge due to the increasing worldwide adoption of LLMs-based virtual chatbot agents and Question-answer systems.  
Although several methods have been recently presented to tackle the problem, some partly work well as countermeasure systems as, at the same time, they may introduce further hallucinations into the LLM itself \cite{varshney2023stitch}.  
Varshney et al. \cite{varshney2023stitch} proposed an effective method to lower GPT3.5 hallucination by 33\%. They addressed hallucinations in generated sentences by instructing the model to rectify them. This involves removing or substituting the false information, supported by retrieved knowledge. 
 
Despite LLMs' hallucination being a relatively new issue, several methods relying on different paradigms have been proposed. They can be grouped into the following families: 
\begin{itemize}
  \item Fine-tuning
  \item Knowledge Graphs
  \item Memory Augmentation
  \item Context Prompts
  \item Preemptive Strategies
\end{itemize}

\begin{figure}
    \centering
\includegraphics[width=\textwidth]{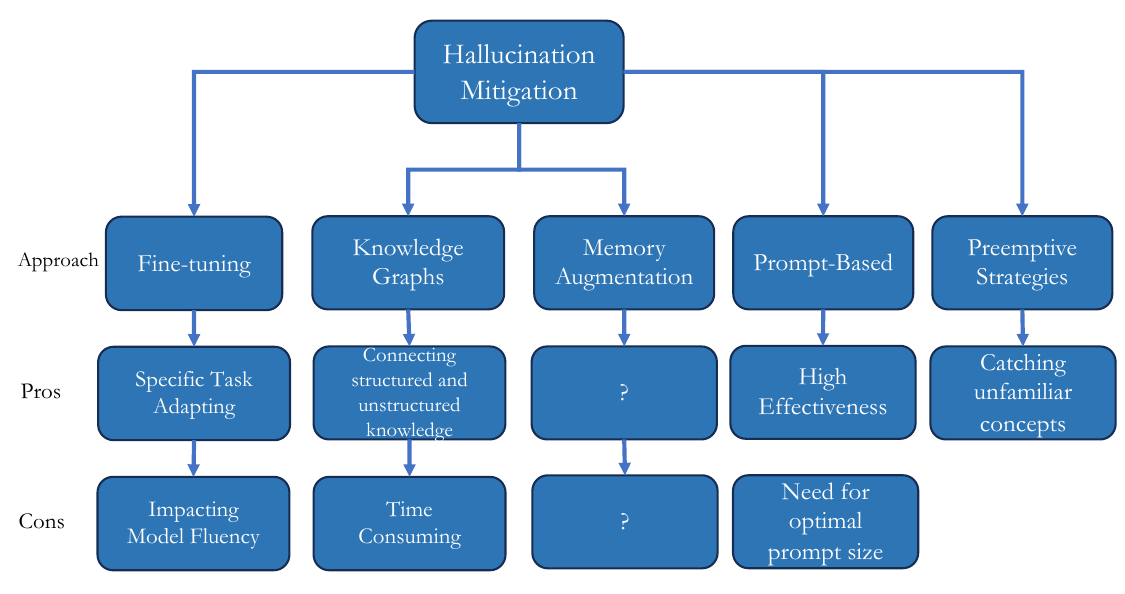}
    \caption{Hallucination Mitigation approaches, pros and cons are depicted above.}
    \label{fig:mitigation}
\end{figure}

A graphical depiction of mitigation methods, pros and cons, is given in Figure \ref{fig:mitigation}. 
Fine-tuning is a well-known technique broadly used in machine learning to specialise a pre-trained model on a specific scenario characterised by a small dataset \cite{church2021emerging}. LLMs' hallucinations can be mitigated with fine-tuning, as Lee et al. showed in their work \cite{lee2019mixout}. However, LLMs featuring millions of parameters make fine-tuning an expensive solution. 
Knowledge graph methods allow for integrating structured and unstructured knowledge \cite{moiseev2022skill}. That gives LLMs a more extended platform to run tasks. The drawback entails two aspects: designing a well-curated knowledge base is time-consuming, and keeping up-to-date knowledge is labour-intensive. 
Wu et al. \cite{wu2022efficient} proposed an augmented transformer for knowledge-intensive NLP tasks. That is due to the need for deep learning methods to extend their capabilities on new knowledge. 
Although NLP models have already benefited from memory augmentation, the same cannot be said for LLMs, as no tests have been run. 

Prompt-based solutions have been recently introduced to 'de-hallucinate' LLMs. 
Jha et al. \cite{jha2023dehallucinating}  proposed a self-monitoring prompting framework. This framework leverages formal methods to identify errors in the LLM's responses autonomously. They employed the conversational abilities of LLMs for response alignment with specified correctness criteria through iterative refinement. 
Luo et al. \cite{luo2023zero} proposed Self-Familiarity, a method to overcome the current SOTA (State-of-the-art) techniques that identify and mitigate hallucinations post-generation. 

Self-Familiarity introduced an innovative zero-resource, pre-detection approach to mitigate the risk of large language models (LLMs) producing inaccurate information. This method extracted and processed conceptual entities from the instruction. Subsequently, it employed prompt engineering to acquire a familiarity score for each concept. These scores were combined to yield the ultimate familiarity score at the instruction level. A low instruction-level familiarity score indicates a higher likelihood of the LLM generating erroneous information, prompting it to abstain from generating a response.

Feldman et al. \cite{feldman2023trapping} designed a method relying on context-tagged prompts. They created a set of questions and then developed context prompts to help the LLM answer those questions more accurately. They then validated the context prompts and the questions to ensure they worked as intended. Finally, they ran experiments with different GPT models to see how context prompts affected the LLM responses' accuracy.

\section{Future Perspective}
Some considerations are drawn in this section concerning LLMs hallucination and mitigation methods.
Zero-resource hallucination detection: Current zero-resource hallucination detection methods are still in their early stages of development. Future research could focus on developing more accurate and reliable methods for a broader range of scenarios.
Black-box hallucination detection: Black-box hallucination detection is even more challenging than zero-resource hallucination detection, as there is no access to the LLM's internal states. Future research could focus on developing new black-box hallucination detection methods or finding ways to make existing methods more effective.
Hallucination detection for specific tasks: Most current hallucination detection methods are general-purpose. However, hallucination detection may be more effective if tailored to specific tasks. For example, hallucination detection methods for factual question answering could be designed to leverage the fact that factually accurate answers are more likely to be grounded in real-world knowledge.
Hallucination detection in multimodal LLMs: Multimodal LLMs are a new type of LLM that can process and generate text, images, and other media types. Hallucination detection in multimodal LLMs is a challenging problem, but it is essential to address, as multimodal LLMs are becoming increasingly popular.
Here are some specific research questions that could be explored in each of these areas:

Zero-resource hallucination detection:
Can zero-resource hallucination detection be made more accurate and reliable?
Can zero-resource hallucination detection be applied to a broader range of scenarios, such as real-time conversation?
Black-box hallucination detection:
Can new methods be developed for black-box hallucination detection?
Can existing hallucination detection methods be made more effective for black-box scenarios?
Hallucination detection for specific tasks:
Can hallucination detection be tailored to specific tasks, such as factual question answering and code generation?
How can we leverage the unique properties of each task to improve the accuracy of hallucination detection?
Hallucination detection in multimodal LLMs:
How can hallucination detection be adapted to multimodal LLMs?
How can we leverage the multimodal capabilities of these models to improve the accuracy of hallucination detection?
In addition to these research questions, developing and evaluating new benchmarks for hallucination detection is also substantial. This will help to ensure that hallucination detection methods are evaluated fairly and consistently.
\begin{acknowledgments}

The contribution is funded by the grant awarded for HeReFaNMi - Health-Related Fake News Mitigation project, selected in the NGI Search 1st Open Call.

\end{acknowledgments}

\bibliography{sample-ceur}

\appendix



\end{document}